\documentclass{article}
\usepackage{microtype}
\usepackage{graphicx}
\usepackage{subfigure}
\usepackage{booktabs}

\usepackage{amsfonts}
\usepackage{amsmath}
\usepackage{amsbsy}
\usepackage [english]{babel}
\usepackage [autostyle, english = american]{csquotes}
\MakeOuterQuote{"}
\usepackage{hyperref}

 \usepackage[accepted]{icml2020}

\icmltitlerunning{Fiedler Regularization: Learning Neural Networks with Graph Sparsity}

\begin{document}

\twocolumn[
\icmltitle{Fiedler Regularization: Learning Neural Networks with Graph Sparsity}
 
 \begin{icmlauthorlist}
\icmlauthor{Edric Tam}{to}
\icmlauthor{David Dunson}{to}

\end{icmlauthorlist}
\icmlaffiliation{to}{Department of Statistical Science, Duke University, Durham, NC, USA}
\icmlcorrespondingauthor{Edric Tam}{edric.tam@duke.edu}

\icmlkeywords{Fiedler Value, Neural Networks, Deep Learning, Regularization, Graph Laplacian}

\vskip 0.3in
]
\printAffiliationsAndNotice{}

\begin{abstract}
We introduce a novel regularization approach for deep learning that incorporates and respects the underlying graphical structure of the neural network. Existing regularization methods often focus on penalizing weights in a global/uniform manner that ignores the connectivity structure of the neural network. We propose to use the Fiedler value of the neural network's underlying graph as a tool for regularization. We provide theoretical support for this approach via spectral graph theory. We show several useful properties of the Fiedler value that make it suitable for regularization. We provide an approximate, variational approach for faster computation during training. We provide an alternative formulation of this framework in the form of a structurally weighted L1 penalty, thus linking our approach to sparsity induction. We performed experiments on datasets that compare Fiedler regularization with traditional regularization methods such as Dropout and weight decay. Results demonstrate the efficacy of Fiedler regularization.
\end{abstract}

\section{Introduction}
Neural networks (NNs) are important tools with many applications in various machine learning domains such as computer vision, natural language processing and reinforcement learning. NNs have been very effective in settings where large labeled datasets are available. Empirical and theoretical evidence has pointed to the ever-increasing capacity of recent NN models, both in depth and width, as an important contributor to their modeling flexibility and success.  However, even the largest datasets can still be potentially overfitted by NNs with millions of parameters or more. A wide range of techniques for regularizing NNs have thus been developed. These techniques often regularize the network from a global/uniform perspective, e.g. weight decay \citep{krogh1992simple}, L1/L2 penalization of weights, dropping nodes/weights in a Bernoulli manner with uniform probability across units/layers \citep{hinton2012improving, srivastava2014dropout,wan2013regularization}, or stopping training early. These commonly used approaches ignore the NN's underlying graphical structure, which can provide valuable connectivity information for regularization. 

One natural generalization of these existing approaches is to take the graph structure of the NN into consideration during regularization. Existing feedforward NN architectures, e.g. multi-layer perceptrons, frequently employ fully connected layers that lead to many redundant paths between nodes of the network. These redundant connections can contribute to over-fitting through the phenomenon of co-adaptation, where weights become dependent on one another, leading to highly correlated behavior amongst different hidden units \citep{hinton2012improving}. Empirical work has shown that dropping weights and nodes randomly during training can significantly improve test performance by reducing co-adaptation \citep{hinton2012improving, srivastava2014dropout,wan2013regularization}. 

In this work, we would like to regularize the NN through reducing co-adaptation and penalizing extraneous connections in a way that respects the NN's graphical/connectivity structure. We introduce Fiedler regularization, borrowing from advances in spectral graph theory \citep{godsil2013algebraic, chung1997spectral, Spielman2019Algebraic}. The Fiedler value of a connected graph, denoted $\lambda_2$, also known as the algebraic connectivity, is the second smallest eigenvalue of the graph's Laplacian matrix. Its magnitude characterizes how well connected a graph is. By adding the Fiedler value as a penalty term to the loss function during training, we can penalize the connectedness of the NN and reduce co-adaptation while taking into account the graph's connectivity structure. We also explore several useful characteristics of the Fiedler value. We show that the Fiedler value is a concave function on the sizes of the NN's weights, implying that using it as a penalty will not substantially worsen the optimization landscape. We additionally show that the Fiedler value's gradient with respect to the network's weights admits a closed form expression, which allows for direct employment of existing gradient-based stochastic optimization techniques for Fiedler regularization.

In practice, for larger networks, to speed up computation, we propose a variational approach, which replaces the original Fiedler value penalty term by a quadratic form of the graph Laplacian. When used together with the so called test vectors, such a Laplacian quadratic form sharply upper bounds the Fiedler value. This variational approximation allows for substantial speedups during training. We give an alternate but equivalent formulation of the variational penalty in terms of a structurally weighted L1 penalization, where the weights depend on the (approximate) second eigenvector of the graph Laplacian. This L1 formulation allows us to link Fiedler regularization to sparsity induction, similar to the parallel literature in statistics \citep{tibshirani1996regression, zou2006adaptive}.

There has been prior work on using the Laplacian structure of the input data to regularize NNs \citep{kipf2016semi, jiang2018graph, zeng2019deep}. There has also been recent work on understanding regularization on NNs in Bayesian \citep{vladimirova2018understanding, polson2018posterior} as well as other settings. These approaches do not consider the graphical connectivity structure of the underlying NN. The main contributions of the paper include: (1) to the authors' best knowledge, this is the first application of spectral graph theory and Fiedler values in regularization of NNs via their own underlying graphical/connectivity structures. (2) We give practical and fast approaches for Fiedler regularization, along with strong theoretical guarantees and experimental performances.

\section{Spectral Graph Theory}
\subsection{Setup and Background}
Let $W$ denote the set of weights of a feedforward NN $\mathbf{f}$. We denote $\mathbf{f}$'s underlying graph structure as $G$. We would like to use structural information from $G$ to regularize $\mathbf{f}$ during training. Feedforward NNs do not allow for self-loops or recurrent connections, hence it suffices that $G$ be a finite, connected, simple, weighted and undirected graph in this setting. Such a graph $G$ can be fully specified by a triplet $(V, E, |W|)$, where $|\cdot|$ denotes the absolute value mapping. The vertex set $V$ of $G$ corresponds to all the units (including units in the input and output layers) in the NN $\mathbf{f}$, while the edge set $E$ of $G$ corresponds to all the edges in $\mathbf{f}$.  For our purposes of regularization, $G$ is restricted to have non-negative weights $|W|$, which are taken to be the absolute value of the corresponding weights $W$ in the NN $\mathbf{f}$. Throughout the paper we will use $n$ to denote the number of vertices in $G$.  $|W|$ and $E$ can be jointly represented by an $n \times n$ weighted adjacency matrix $\mathbf{|W|}$, where $\mathbf{|W|}_{ij}$ is the weight on the edge $(i,j)$ if vertices $i$ and $j$ are connected, and 0 otherwise. The degree matrix $\mathbf{D}$ of the graph is a $n \times n$ diagonal matrix, where $\mathbf{D}_{ii} = \sum_{j = 1}^n \mathbf{|W|}_{ij}$. The Laplacian matrix $\mathbf{L}$, which is a central object of study in spectral graph theory, is defined as the difference between the degree and the adjacency matrix, i.e. $\mathbf{L} = \mathbf{D}-\mathbf{|W|}$. In certain contexts where we would like to emphasize the dependency of $\mathbf{L}$ on the particular graph $G$ or the weights $|\mathbf{W}|$, we will adopt the notation $\mathbf{L}_G$ or $\mathbf{L}_{|\mathbf{W}|}$. We use $[M]$ to denote the set $\{1,2,\cdots, M\}$ for positive integer $M$. Throughout the paper, eigenvalues are real-valued since the matrices under consideration are symmetric. We adopt the convention where all eigenvectors are taken to be unit vectors. We order the eigenvalues in ascending order, so $\lambda_i \leq \lambda_j$ for $i < j$. When we want to emphasize $\lambda_i$ as a function of the weights, we use $\lambda_i(\mathbf{|W|})$. We use $\mathbf{v}_i$ to denote the corresponding eigenvector for $\lambda_i$. We use $n(S)$ to denote the cardinality of a given set $S$. 

\subsection{Graph Laplacian}
The Laplacian matrix encodes much information about the structure of a graph. One particularly useful characterization of the graph Laplacian is through the so called Laplacian quadratic form \citep{batson2012twice, Spielman2019Algebraic, chung1997spectral}, defined below.

{\bf Definition 2.2.1 (Laplacian quadratic form)} {\it Given a graph $G = (V, E, |W|)$, define its Laplacian quadratic form $Q_G: \mathbb{R}^{|V|} \to \mathbb{R}^+$ as
\[
	Q_G(\textbf{z}) := \textbf{z}^T \mathbf{L}_G \textbf{z} = \sum_{(i,j) \in E} \mathbf{|W|}_{ij} (\textbf{z}(i) - \textbf{z}(j))^2
\]
where $\textbf{z}(k)$ denotes the k\textsuperscript{th} entry of the vector $\textbf{z} \in \mathbb{R}^{|V|}$.
} 

The Laplacian quadratic form demonstrates how a graph's boundary information can be recovered from its Laplacian matrix. One can think of $\textbf{z}$ as a mapping that assigns to each vertex a value. If we denote the characteristic vector of a subset of vertices $S \subset V$ as $\textbf{1}_S$, i.e. $\textbf{1}_S (i) = 1$ if $i \in S$ and $\textbf{1}_S (i) = 0$ otherwise, and apply it to the Laplacian quadratic form, we obtain $\textbf{1}_S^T \mathbf{L}_G \textbf{1}_S = \sum_{(i,j) \in E, i\in S, j \not \in S} \mathbf{|W|}_{ij}$. This expression characterizes the size of the graph cut $(S,  V-S)$, which is the sum of the weights of edges crossing the boundary between $S$ and $V-S$. The size of any graph cut can therefore be obtained by application of the corresponding characteristic vector on the Laplacian quadratic form.  

\subsection{Edge Expansion and Cheeger's Inequality}
The above discussion on the sizes of graph cuts is highly related to our study of regularizing a NN. Reducing the sizes of graph cuts in a NN would imply reducing the NN's connectivity and potentially co-adaptation. A related but more convenient construct that captures this notion of boundary sizes in a graph is the graph's edge expansion (also known as the Cheeger constant or the isoperimetric number), which can be informally thought of as the smallest "surface-area-to-volume ratio" achieved by a subset of vertices not exceeding half of the graph.

{\bf Definition 2.3.1 (Edge expansion of a graph)} {\it The edge expansion $\phi_G$ of a graph $G = (V, E, |W|)$ is defined as 
\[
\phi_G = \min_{S \subset V, n(S) \leq \frac{n(V)}{2}} \frac{\sum_{i \in S, j \not \in S} \mathbf{|W|}_{ij}}{n(S)},
\]
where $n(S)$ denotes the number of vertices in $S$. 
}  

Observe that the term in the numerator characterizes the size of the graph cut $(S, V-S)$, while the denominator normalizes the expression by the number of vertices in $S$. The edge expansion is then taken to be the smallest such ratio achieved by a set of vertices $S$ that has cardinality at most half that of $V$. One can think of the edge expansion of a graph as characterizing the connectivity bottleneck of a graph. It is highly related to how sparse the graph is and whether there exist nice planar embeddings of the graph \citep{hall1970r}.  

We would like to control the edge expansion of the NN's underlying graph for regularization. Direct optimization of edge expansion is a difficult problem due to the combinatorial structure. Instead, we control the edge expansion indirectly through the Fiedler value $\lambda_2$, the second smallest eigenvalue of the graph's Laplacian $\mathbf{L}$.  $\lambda_2$ is related to the edge expansion of the graph through Cheeger's inequality \citep{Spielman2019Algebraic, chung1997spectral, godsil2013algebraic}.

{\bf Proposition 2.3.2 (Cheeger's inequality) } {\it Given a graph $G = (V, E, |W|)$, the edge expansion $\phi_G$ is upper and lower bounded as follows: 
\[ 
	\sqrt{2d_{\max}(G)\lambda_2}\geq \phi_G \geq \frac{\lambda_2}{2},
\]
where $d_{\max}(G)$ is the maximum (weighted) degree of vertices in $G$ and $\lambda_2$, the Fiedler value, is the second smallest eigenvalue of $G$'s Laplacian matrix. 
} 

Cheeger's inequality originally arose from the study of Riemannian manifolds and was later extended to graphs. There are many versions of Cheeger's inequality depending on the types of graphs and normalizations used. The proofs can be found in many references, including \citep{Spielman2019Algebraic, chung1997spectral, godsil2013algebraic}. These types of Cheeger's inequalities are generally tight asymptotically up to constant factors.

The Fiedler value is also known as the algebraic connectivity because it encodes considerable information about the connectedness of a graph. Cheeger's inequality allows sharp control over the edge expansion of $G$ via the Fiedler value. By making the Fiedler value small, we can force the edge expansion of the graph to be small, thus reducing the connectedness of the graph and potentially alleviating co-adaptation in the NN setting. On the other hand, a large Fiedler value necessarily implies a large edge expansion. This shows that penalization of the Fiedler value during NN training is a promising regularization strategy to reduce connectivity and thus co-adaptation.

\section{Fiedler Regularization}

\subsection{Supervised Classification Setup}
We now consider the classical setup for classification. Given training data $\{\mathbf{x}_i, y_i\}_{i = 1}^N$, where $i$ indexes the $N$ data points, $\mathbf{x} \in \mathbb{R}^d$ denotes the independent $d$-dimensional feature vectors, $y \in [M]$ denotes the labels and $M$ is the number of categories, we aim to find a mapping $\mathbf{f}$ that predicts $y$ through $\mathbf{f}(\mathbf{x})$ so that some pre-specified loss $\mathcal{L}(f(\mathbf{x}), y)$ is minimized in the test data. Typical choices of the loss function $\mathcal{L}(\cdot,\cdot)$ include the cross-entropy loss, hinge loss etc. It is assumed that new observations in the testing set follow the same distribution as the training data. We will denote the estimator of $\mathbf{f}$ as $ \mathbf{\hat f}$. When we want to emphasize the estimator's dependence on the weights, we will use $ \mathbf{ \hat f_W}$ .

For feedforward NNs, $\mathbf{\hat f}$ is characterized as a composition of non-linear functions $\{ \mathbf{g}^{(l)}\}_{l = 1}^\Lambda$, i.e. $\mathbf{\hat f}= \mathbf{g}^{(\Lambda)} \circ \mathbf{g}^{(\Lambda-1)} \circ \cdots \circ \mathbf{g}^{(1)}(\mathbf{x})$, where $\Lambda$ is the number of layers in the network. The outputs of the $l$\textsuperscript{th} layer have the form $\mathbf{h}^{(l)} := \mathbf{g}^{(l)}(\mathbf{h}^{(l-1)}) := \pmb{\sigma}^{(l)}(\mathbf{W}^{(l)} \mathbf{h}^{(l-1)} + \mathbf{b}^{(l)})$, where $\pmb{\sigma}^{(l)}$, $\mathbf{W}^{(l)}$ and $\mathbf{b}^{(l)}$ are the activation function, weight matrix and bias of the $l$\textsuperscript{th} layer of the NN, respectively. In essence, each hidden layer first performs an affine transformation on the previous layer's outputs, followed by an element-wise activation that is generally nonlinear. For more details on this setup, see the excellent review \citep{fan2019selective}.

\subsection{Penalizing with Fiedler Value}
 Given the motivations from section 2, we would like to penalize the connectivity of the NN during training. In the Fiedler regularization approach, we add $\lambda_2$ as a penalty term to the objective.

{\bf Definition 3.2.1 (Fiedler regularization)} {\it During training of the neural network, we optimize the following objective:
\[
	\min_{\mathbf{W}} \mathcal{L}(Y, \mathbf{\hat{f}_W}(X)) + \delta \lambda_2(\mathbf{|W|}),
\]
where $\lambda_2(\mathbf{|W|})$ is the Fiedler value of the NN's underlying graph, $\mathbf{W}$ is the weight matrix of the NN, $\delta$ is a tuning parameter, and $Y$ and $X$ denote the training labels and training features, respectively. 
}

We remark that the actual NN $\mathbf{\hat f_W}$ will have weights that can be negative, but the regularization term $\lambda_2(\mathbf{|W|})$ will only depend on the sizes of such weights, which can be thought of as edge capacities of the underlying graph $G$. 

Note that one can generalize the penalty by applying a differentiable function $Q$ to $\lambda_2$. For our discussion below, we will focus on the case where no $Q$ is applied in order to focus the analysis on the Fiedler value. However, incorporating $Q$ is straightforward, and desirable properties such as having a closed-form gradient can generally be retained. We also note that, without loss of generality, one can consider the biases of the units in the NN as additional weights with constant inputs, so it is straightforward to include consideration of both biases and weights in Fiedler regularization.

For our purposes, we consider the main parameters of interest to be the weights of the NN. The choice of the activation function(s), architecture, as well as the choice of hyperparameters such as the learning rate or the tuning factor etc. are all considered to be pre-specified in our study. There is a separate and rich literature devoted to methods for selecting activation functions/architectures/hyperparameters that we will not consider here. 

\subsection{Properties of the Fiedler Value}

It is instructive to examine some properties of the Fiedler value in order to understand why it is an appropriate tool for regularization.

First, one concern is whether using the Fiedler value as a penalty in the objective would complicate the optimization process. The Fiedler value can be viewed as a root of the Laplacian matrix's characteristic polynomial, which in higher dimensions has no closed-form solution and can depend on the network's weights in a convoluted manner.

To address this concern, the following proposition shows that the Fiedler penalty is a concave function of the sizes of the NN's weights. This shows that when we add the Fiedler penalty to deep learning objectives, which are typically highly non-convex, we are not adding substantially to the optimization problem's difficulty.

{\bf Proposition 3.3.1 (Concavity of Fiedler Value)} {\it The function $\lambda_2(|W|)$ 
is a concave function of the sizes of the NN's weights $|W|$.

} 
{\bf Proof:}{ Since the Fiedler value $\lambda_2$ is just the second smallest eigenvalue of the Laplacian, and we know that the first eigenvector of the Laplacian must be constant, we can consider $\lambda_2$'s Rayleigh-Ritz variational characterization as follows: 
$$\lambda_2(\mathbf{|W|}) = \inf_{||\textbf{u}|| = 1, \textbf{u}^T\textbf{1} = 0} \textbf{u}^T\mathbf{L}\textbf{u}$$
$$= \inf_{||\textbf{u}|| = 1, \textbf{u}^T\textbf{1} = 0} \sum_{(i,j) \in E} \mathbf{|W|}_{ij}(\textbf{u}(i) - \textbf{u}(j))^2$$ 
Note that this is a pointwise infimum of a linear function of $\mathbf{|W|}_{ij}$. Since linear functions are concave (and convex), and the pointwise infimum preserves concavity, we have that $\lambda_2$ is a concave function of the sizes of the weights. $\square$

}

This is related to Laplacian eigenvalue optimization problems and we refer to \citep{boyd2006convex} and \citep{sun2006fastest} for a more general treatment.

\subsection{Closed-form Expression of Gradient}
In all except the most simple of cases, optimizing the loss function $\text{min}_{\mathbf{W}} \mathcal{L}(Y, \mathbf{f_W}(X))$ is a non-convex problem. There are a variety of scalable, stochastic algorithms for practical optimization on such objectives. Virtually all of the widely used methods, such as stochastic gradient descent (SGD) \citep{ruder2016overview}, Adam \citep{kingma2014adam}, Adagrad \citep{duchi2011adaptive}, RMSProp \citep{graves2013generating} etc, require computation of the gradient of the objective with respect to the parameters. We provide a closed-form analytical expression of the gradients of a general Laplacian eigenvalue with respect to the entries of the Laplacian matrix. From that, as a straightforward corollary, a closed-form analytical expression of the Fiedler value's gradient is obtained.

{\bf Proposition 3.4.1 (Gradient of Laplacian Eigenvalue) } {\it Assuming that the eigenvalues of the Laplacian $\mathbf{L}$ are not repeated, the gradient of the $k$\textsuperscript{th} smallest eigenvalue $\lambda_k$ with respect to $\mathbf{L}$'s $(ij)$\textsuperscript{th} entry $\mathbf{L}_{ij}$ can be analytically expressed as
$$\frac{d\lambda_k}{d\mathbf{L}_{ij}} = \textbf{v}_k(i)\times \textbf{v}_k(j),$$
where $\textbf{v}_k(i)$ denotes the $i$\textsuperscript{th} entry of the $k$\textsuperscript{th} eigenvector of the Laplacian. We adopt the convention in which all eigenvectors under consideration are unit vectors. 
}

{\bf Proof:}
{To compute $\frac{d\lambda_k}{d\mathbf{L}_{ij}}$, note that since the Laplacian matrix is symmetric, all eigenvalues are real. By assumption, the eigenvalues are not repeated. Under this situation, there is an existing closed-form formulae (see \citep{petersen2008matrix}): $\textbf{v}_k^T(\partial \mathbf{L})\textbf{v}_k = d\lambda_k$. Specializing to individual entries, we get $\frac{d\lambda_k}{d\mathbf{L}_{ij}} = \textbf{v}_k(i)\times \textbf{v}_k(j)$. $\square$
}

{\bf Corollary 3.4.2 (Gradient of Fiedler value with respect to weights) } {\it As an immediate special case of Proposition 3.4.1, the gradient of the Fiedler value $\lambda_2$ can be expressed as: 
$$\frac{d\lambda_2}{d\mathbf{L}_{ij}} = \textbf{v}_2(i) \times \textbf{v}_2(j)$$
Since $\mathbf{L}_{ij} = -\mathbf{|W|}_{ij}$ for $i \not = j$, this yields the following gradient for non-zero weights: 
$$\frac{d\lambda_2}{d\mathbf{\mathbf{|W|}}_{ij}} = -\textbf{v}_2(i) \times \textbf{v}_2(j)$$
}
\\One can remove the absolute value mapping in the gradient by a simple application of the chain rule. Using the above gradient expression, we can perform weight updates in existing deep learning libraries using a wide variety of stochastic optimization methods.  

Our assumption that the eigenvalues are not repeated generally holds in the context of NNs. The weights in a NN are usually initialized as independent draws from certain continuous distributions, such as the uniform or the Gaussian. Repeated Laplacian eigenvalues often occur when there are strong symmetries in the graph. Such symmetries are typically broken in the context of NNs since the probability of different weights taking the same non-zero value at initialization or during training is negligible.

\section{Variational, Approximate Approach for Computational Speedup}
We have given theoretical motivation and outlined the use of the Fiedler value as a tool for NN regularization. However, typical matrix computations for eigenvalues and eigenvectors are of order $O(n^3)$, where $n = |V|$. This can be computationally prohibitive for moderate to large networks. Even though there exist theoretically much more efficient algorithms to approximate eigenvalues/eigenvectors of graph-related matrices, in practice computing the Fiedler value in every iteration of training can prove costly. To circumvent this issue, we propose an approximate, variational approach to speed up the computation to $O(|E|)$, where $|E|$ is the number of edges in the graph. This proposed approach can be readily implemented in popular deep learning packages such as Tensorflow and PyTorch.

We make the following observations. First, for the purpose of regularizing a NN, we do not need the exact Fiedler value of the Laplacian matrix. A good approximation suffices. Second, there is no need to update our approximate $\lambda_2$ at every iteration during training. We can set a schedule to update our $\lambda_2$ approximation periodically, say once every $100$ iterations. The frequency of updates can be treated as a new hyperparameter. Both of these observations allow for substantial speedups in practice. An outline of the pseudo-code for this approximate, variational approach is provided in Algorithm 1 below. 

To obtain an approximate Fiedler value, we use a special type of Laplacian quadratic form involving the so called test vectors.
We additionally provide a perturbation bound that gives justification for periodically updating the Fiedler value.

\subsection{Rayleigh Quotient Characterization of Eigenvalues}
We can closely upper-bound the Fiedler value via the notion of test vectors \citep{Spielman2019Algebraic}, which depends crucially upon the Rayleigh quotient characterization of eigenvalues. 

{\bf Proposition 4.1.1 (Test Vector Bound) } {\it For any unit vector $\textbf{u}$ that is perpendicular to the constant vector $\textbf{1}$, we have: $$\lambda_2 \leq \textbf{u}^T\mathbf{L}\textbf{u}$$  Any such unit vector is called a test vector. Equality is achieved when $\textbf{u} = \textbf{v}_2$.
} 

{\bf Proof:}
{The Laplacian matrix of a non-negatively weighted graph is symmetric and positive semidefinite. Thus all eigenvalues are real and non-negative. In particular, the smallest eigenvalue of the Laplacian is 0, with the constant vector being the first eigenvector. For a connected graph, the second smallest eigenvalue, which is the Fiedler value, can thus be variationally characterized by $\lambda_2 = \min_{||\textbf{u}|| = 1, \textbf{u}^T \textbf{1} = 0}\textbf{u}^T\mathbf{L}\textbf{u}$. This gives us the desired upper bound.} $\square$

The above description implies that by appropriately choosing test vectors, we can effectively upper bound the Fiedler value. This in turn implies that during training, instead of penalizing by the exact Fiedler value, we can penalize by the quadratic form upper bound instead.	
In other words, we would perform the following optimization,
$$\min_{\mathbf{W}} \mathcal{L}(Y, \mathbf{\hat f_W}(X)) + \delta \mathbf{u}^T\mathbf{L}\mathbf{u}$$
For a given $\mathbf{u}$, this speeds up computation of the penalty term considerably to $O(|E|)$ since $\textbf{u}^T\mathbf{L}\textbf{u} = \sum_{(i,j)\in E}\mathbf{|W|}_{ij}(\textbf{u}(i) - \textbf{u}(j))^2$. The core question now becomes how to choose appropriate test vectors $\mathbf{u}$ that are close to $\textbf{v}_2$. 

We propose to initialize training with the exact $\textbf{v}_2$ and recompute/update $\textbf{v}_2$ only periodically during training. For an iteration in between two exact $\textbf{v}_2$ updates, the $\textbf{v}_2$ from the previous update carries over and serves as the test vector for the current iteration. By proposition 4.1.1, this will always upper-bound the true $\lambda_2$.

During training, weights of the NN are updated at each iteration. This is equivalent to adding a symmetric matrix $\mathbf{H}$ to the Laplacian matrix $\mathbf{L}$, which is also symmetric, at every iteration. We can bound the effect of such a perturbation on the eigenvalues via Weyl's inequality \citep{horn2012matrix, horn1998eigenvalue}. 

{\bf Proposition 4.1.2 (Weyl's Inequality)} {\it 
Given a symmetric matrix $\mathbf{L}$ and a symmetric perturbation matrix $\mathbf{H}$, both with dimension $n \times n$, for any $1\leq i \leq n$, we have: 
$$|\lambda_i(\mathbf{L}+\mathbf{H}) - \lambda_i(\mathbf{L})| \leq ||\mathbf{H}||_{op}$$
where $||\cdot||_{op}$ denotes the operator norm. 
} 

Weyl's inequality is a classic result in matrix analysis, and its proof can be found in the excellent reference \citep{horn2012matrix} as a straightforward result of linearity and the Courant-Fischer theorem. As an immediate special case, $|\lambda_2(\mathbf{L}+\mathbf{H}) - \lambda_2(\mathbf{L})| \leq ||\mathbf{H}||_{op}$. Proposition 4.1.2 tells us that as long as the perturbation $\mathbf{H}$ is small, the change in Fiedler value caused by the perturbation will also be small. In fact, this shows that the map $\mathbf{L}\to \lambda_2(\mathbf{L})$ is Lipschitz continuous on the space of symmetric matrices. In the context of training NNs, $\mathbf{H}$ represents the updates to the weights of the network during training. This justifies updating $\lambda_2$ only periodically for the purposes of regularization. It suggests that with a smaller learning rate, $\mathbf{H}$ would be smaller, and therefore the change to $\lambda_2$ would also be smaller, and updates of the test vectors can be more spaced apart. On the other hand, for larger learning rates we recommend using updating test vectors more frequently. Alternatively, a similar bound on eigenvector perturbation could be established via the Davis-Kahan Sin-$\Theta$ inequality, but is omitted here since the eigenvalue bound suffices for Fiedler regularization. 

\begin{algorithm}[tb]
   \caption{Variational Fiedler Regularization with SGD}
   \label{alg:fiedler}
\begin{algorithmic}
   \STATE {\bf Input:} Training data $\{\mathbf{x}_i, y_i\}_{i = 1}^N$
   \STATE {\bf Hyperparameters:} Learning rate $\eta$, batch size $m$, penalty parameter $\delta$, updating period $T$
   \STATE {\bf Algorithm:}
   \STATE Initialize parameters $\mathbf{W}$ of the NN
   \STATE Compute the Laplacian $\mathbf{L_{|W|}}$ of the NN
   \STATE Compute the Fiedler vector $\textbf{v}_2$ of the Laplacian $\mathbf{L_{|W|}}$
   \STATE Set $\textbf{u} \leftarrow \textbf{v}_2$
   \STATE Initialize counter $c = 0$ 
   \WHILE{Stopping criterion not met}
   \STATE Sample minibatch $\{\mathbf{x}^{(i)}, y^{(i)}\}_{i = 1}^m$ from training set 
   \STATE Set gradient $\pmb{\gamma} = \textbf{0}$
   \FOR{ $i=1$ to $m$}
   \STATE Compute gradient 
   $\pmb{\gamma} \leftarrow \pmb{\gamma} + \nabla_{\mathbf{W}} \mathcal{L}(\mathbf{\hat{f}_W}(\mathbf{x}^{(i)}), y^{(i)}) + \delta \nabla_{\mathbf{W}}\textbf{u}^T \mathbf{L_{|W|}} \textbf{u}$ 
   \ENDFOR
   \STATE Apply gradient update  $\mathbf{W} \leftarrow \mathbf{W} - \eta \pmb{\gamma}$
   \STATE Update Laplacian matrix $\mathbf{L_{|W|}}$
   \STATE Update counter $c \leftarrow c + 1$
   \IF{$c$ \text{mod} $T = 0$}
   \STATE Recompute $\textbf{v}_2$ from $\mathbf{L_{|W|}}$
   \STATE Set $\textbf{u} \leftarrow \textbf{v}_2$
   \ENDIF
   \ENDWHILE

\end{algorithmic}
\end{algorithm}

\section{Weighted-L1 formulation and Sparsity}
We now expand on an equivalent formulation of the variational Fiedler penalty as a weighted L1 penalty. We note that the Laplacian quadratic form in the variational Fiedler penalty can be written as $ \textbf{u}^T\textbf{L}\textbf{u} = \sum_{(i,j) \in E} \mathbf{|W|}_{ij} (\textbf{u}(i) - \textbf{u}(j))^2$. This yields the variational objective: 
$$ \min_{\mathbf{W}} \mathcal{L}(Y, \mathbf{\hat{f}_W}(X)) + \delta \sum_{(i,j) \in E} \mathbf{|W|}_{ij} (\textbf{u}(i) - \textbf{u}(j))^2$$ 
We note that both $\mathbf{|W|}_{ij}$ and $(\textbf{u}(i) - \textbf{u}(j))^2$ are non-negative.  This is equivalent to performing $L_1$ penalization on $\mathbf{W}_{ij}$ with weights $(\textbf{u}(i) - \textbf{u}(j))^2$. 

There is an immense literature on modified $L_1$ penalties in shallow models \citep{zou2006adaptive, candes2008enhancing}. It is well known that optimizing an objective under (weighted) $L_1$ constraints often yields sparse solutions. This thus connects our Fiedler regularization approach with sparsity induction on the weights of the NN. 

In Fiedler regularization, the weights $\mathbf{|W|}_{ij}$ are scaled by a factor of $(\textbf{u}(i) - \textbf{u}(j))^2$, where $\textbf{u}$ is a test vector that approximates $\textbf{v}_2$. From the spectral clustering literature \citep{hagen1992new, donath1972algorithms}, we understand that $\textbf{v}_2$ is very useful for approximating the minimum conductance cut of a graph. The usual heuristic is that one sorts entries of $\textbf{v}_2$ in ascending order, sets a threshold $t$, and groups all vertices $i$ having $\textbf{v}_2 > t$ into one cluster and the rest into another cluster. If the threshold $t$ is chosen optimally, this clustering will be a good approximation to the minimum conductance cut of a graph. As such, the farther apart $\textbf{v}_2(i)$ and $\textbf{v}_2(j)$ are, (1) the edge between nodes $i$ and $j$ (if it exists) is likely "less important" for the connectivity structure of the graph (2) the more likely that nodes $i$ and $j$ belong to different clusters. 

This is also connected to spectral drawing of graphs \citep{Spielman2019Algebraic, hall1970r}, where it can be shown that "nice" planar embeddings/drawings of the graph do not exist if $\lambda_2$ is large. In this sense, Fiedler regularization is forcing the NN to be "more planar" while respecting its connectivity structure. 

Hence, with Fiedler regularization, the penalization on $\mathbf{|W|}_{ij}$ is the strongest when this edge is "less important" for the graph's connectivity structure. This would in theory lead to greater sparsity 
in edges that have low weights and connect distant vertices belonging to different clusters. Thus, Fiedler regularization sparsifies the NN in a way that respects its connectivity structure.

To this end, an alternative guarantee is the ordering property of Laplacian eigenvalues with respect to a sequence of graphs \citep{godsil2013algebraic}.

{\bf Proposition 5.1 (Laplacian Eigenvalue Ordering) } {\it Given the graph $G$ with non-negative weights, if we remove an edge $(a,b)$ to obtain $G\setminus (a,b)$, we have that $$\lambda_2(G\setminus (a,b)) \leq \lambda_2(G)$$
} 
{\bf Proof:}
{
The proof is a simple generalization of \citep{godsil2013algebraic}. Pick $\mathbf{q}_2$ and $\mathbf{r}_2$ to be second eigenvectors of $G$ and $G\setminus(a,b)$ respectively. By the Laplacian quadratic form characterization of eigenvalues, we have $$\lambda_2(G) = \mathbf{q}_2^T\mathbf{L}_G\mathbf{q}_2 = \sum_{(c,d) \in E} \mathbf{|W|}_{cd}(\mathbf{q}_2(c) - \mathbf{q}_2(d))^2$$ $$ \geq [\sum_{(c,d) \in E} \mathbf{|W|}_{cd}(\mathbf{q}_2(c) - \mathbf{q}_2(d))^2] - \mathbf{|W|}_{ab}(\mathbf{q}_2(a) - \mathbf{q}_2(b))^2$$ $$ = \mathbf{q}_2^T\mathbf{L}_{G\setminus (a,b)}\mathbf{q}_2 \geq \mathbf{r}_2^T\mathbf{L}_{G\setminus (a,b)}\mathbf{r}_2 = \lambda_2(G\setminus (a,b))$$ 
where the first inequality follows from the non-negativity of the weights under consideration and the second inequality follows from the Rayleigh-Ritz variational characterization of eigenvalues. 
} $\square$

Hence, by sparsifying edges and reducing edge weights, we are in effect reducing the Fiedler value of the NN. The above ordering property is a special case of the more general Laplacian eigenvalue interlacing property, where $\lambda_2(G\setminus (a,b))$ also admits a corresponding lower bound. For a general treatment, see \citep{godsil2013algebraic}.

We remark that since Fiedler regularization encourages sparsity, during the training process the NN might become disconnected, rendering $\lambda_2$ to become 0. This issue could be easily avoided in practice by dropping one of the disconnected components (e.g. the smaller one) from the Laplacian matrix during training, i.e. remove the Laplacian matrix's rows and columns that correspond to the vertices in the dropped component.

\section{Experiments and Results}
Deep/multilayer feedforward NNs are useful in many classification problems, ranging from popular image recognition tasks to scientific and biomedical problems such as classifying diseases.  We examine the performance of Fiedler regularization on the standard benchmark image classification datasets MNIST and CIFAR10. We also tested Fiedler regularization on the TCGA RNA-Seq PANCAN tumor classification dataset \citep{weinstein2013cancer} from the UCI Machine Learning Repository. We compare the performance of several standard regularization approaches for NNs on these datasets, including Dropout, L1 regularization and weight decay. Extension of such experiments to other classification tasks is straightforward. 

The purpose of the experiments below is not to use the deepest NNs, the latest architectures or the most optimized hyperparameters. Nor is the purpose to show the supremacy of NNs versus other classification methods like random forests or logistic regression. Rather, we attempt to compare the efficacy of Fiedler regularization against other NN regularization techniques as a proof of concept. Extensions to more complicated and general network architectures are explored in the discussion section.

For all our experiments, we consider 5-layer feedforward NNs with ReLU activations and fully connected layers. We used PyTorch 1.4 and Python 3.6 for all experiments. For optimization, we adopted stochastic gradient descent with a momentum of 0.9 for optimization and a learning rate of 0.001. To select the dropping probability for Dropout, as well as the regularization hyperparameter for L1, Fiedler regularization and weight decay, we performed a very rough grid search on a small validation dataset. The Dropout probability is selected to be 0.5 for all layers, and the regularization hyperparameters for L1, Fiedler regularization and weight decay are 0.001, 0.01 and 0.01 respectively. All models in the experiments were trained under the cross-entropy loss. Each experiment was run 5 times, with the median and the standard deviation of the performances reported. All experiments were run on a Unix machine with an Intel Core i7 processor. The code used for the experiments could be found at the first author's Github repository (\href{https://github.com/edrictam/FiedlerRegularization}{https://github.com/edrictam/FiedlerRegularization}).

\subsection{MNIST}
\textbf{Dataset and setup} MNIST is a standard handwriting recognition dataset that consists of 60,000 $28 \times 28$ training images of individual hand-written digits and $10,000$ testing images. We picked the hidden layers of our NN to be 500 units wide. We used a batch size of 100 and the networks were trained with 10 epochs.

\textbf{Results} The results for MNIST are displayed in Table 1. For the MNIST dataset, we obtained very good accuracies for all the methods, with Fiedler regularization standing out, followed by weight decay and Dropout. The high accuracies obtained for the MNIST dataset with feedforward NNs are consistent with results from previous studies \citep{srivastava2014dropout}. Fiedler regularization showed gains over its competitors. 

\begin{table}[t]
\caption{Classification accuracies for MNIST under various regularization schemes (units in percentages)}
\label{sample-table-1}
\vskip 0.15in
\begin{center}
\begin{small}
\begin{sc}
\begin{tabular}{lccr}
\toprule
Regularization & Training & Testing  \\
\midrule
L1   & 90.32  $\pm$ 0.17& 90.25$\pm$ 0.35\\
Weight Decay & 95.12$\pm$ 0.06& 94.98$\pm$ 0.07\\
Dropout & 94.52$\pm$ 0.07 &94.5$\pm$ 0.18 \\
Fiedler & 96.54$\pm$ 0.08 &96.1$\pm$ 0.12 \\
\bottomrule
\end{tabular}
\end{sc}
\end{small}
\end{center}
\vskip -0.1in
\end{table}

\subsection{CIFAR10}
\textbf{Dataset and setup}
CIFAR10 is a benchmark object recognition dataset that consists of $32\times32\times 3$ down-sampled RGB color
images of 10 different object classes. There are 50,000 training images and 10,000 test images in the dataset. We picked the hidden layers of our NN to be 500 units wide. We used a batch size of 100 and the networks were trained with 10 epochs.

\textbf{Results} The results for CIFAR10 are displayed in Table 2. While this is a more difficult image classification task than MNIST, the ordering of performances among the regularization methods is similar. Fiedler regularization performed the best, followed by Dropout, weight decay and L1 regularization. 

\begin{table}[t]
\caption{Classification accuracies for CIFAR10 under various regularization schemes (units in percentages)}
\label{sample-table-2}
\vskip 0.15in
\begin{center}
\begin{small}
\begin{sc}
\begin{tabular}{lccr}
\toprule
Regularization & Training & Testing \\
\midrule
L1   & 28.52$\pm$ 0.9& 28.88$\pm$ 1.11\\
Weight Decay  & 52.93 $\pm$ 0.17& 50.55$\pm$ 0.39\\
Dropout & 46.61$\pm$ 0.35& 44.63$\pm$ 0.39 \\
Fiedler & 57.99 $\pm$ 0.13 & 52.26$\pm$ 0.27 \\
\bottomrule
\end{tabular}
\end{sc}
\end{small}
\end{center}
\vskip -0.1in
\end{table}

\subsection{TCGA Cancer Classification}
\textbf{Dataset and setup} The TCGA Pan-Cancer tumor classification dataset consists of RNA-sequencing as well as cancer classification results for 800 subjects. The input features are 20531-dimensional vectors of gene expression levels, whereas the outputs are tumor classification labels (there are 5 different tumor types under consideration). We used 600 subjects for training and 200 for testing. Due to the highly over-parametrized nature of this classification task, We picked the width of the hidden layers to be narrower, at 50 units. We used a batch size of 10 and the networks were trained with 5 epochs.

\textbf{Results} 
The results for the TCGA tumor classification experiment are displayed in Table 3. Fiedler Regularization and L1 had similarly high performances, followed by weight decay. It is interesting that Dropout achieved a relatively low accuracy, slightly better than chance. Note that since this dataset is relatively small (the testing set has only 200 data points and training set 600), the standard deviation of the accuracies are higher. Notice here that L1 and Fiedler regularization, which explicitly induce sparsity, performed the best. 
\begin{table}[t]
\caption{Classification accuracies for TCGA under various regularization schemes (units in percentages)}
\label{sample-table-3}
\vskip 0.15in
\begin{center}
\begin{small}
\begin{sc}
\begin{tabular}{lccr}
\toprule
Regularization & Training & Testing  \\
\midrule
L1   & 91.5  $\pm$ 13.73& 94.53$\pm$ 12.46\\
Weight Decay & 57$\pm$ 22.42& 60.2$\pm$ 20.94\\
Dropout & 25.33$\pm$ 5.9 &23.88$\pm$ 7.36 \\
Fiedler & 93.33$\pm$ 20.31 &90.55$\pm$ 20.73 \\
\bottomrule
\end{tabular}
\end{sc}
\end{small}
\end{center}
\vskip -0.1in
\end{table}

\subsection{Analysis of Results}

We note that both MNIST and CIFAR10 have more training samples than the dimension of their features. The results from MNIST and CIFAR10 largely agree with each other and confirm the efficacy of Fiedler regularization. Under this setting, other regularization methods like Dropout and weight decay also exhibited decent performance. 

On the other hand, in the TCGA dataset, where the dimension of the input features (20531) is much higher than the number of training samples (600), L1 and Fiedler regularization, which explicitly induce sparsity, performed substantially better than Dropout and weight decay. An inspection of the TCGA dataset suggests that many of the gene expression levels in the input features are 0 (suggesting non-expression of genes), which likely implies that many of the weights in the network, particularly at the input layer, are not essential.  It is thus not surprising that regularization methods that explicitly induce sparsity performs better in these "large p, small n" scenarios, often found in biomedical applications. 

We remark that Fiedler regularization enjoys practical running speeds that are fast, generally comparable to (but slightly slower than) that of most commonly used regularization schemes such as L1 and Dropout. The running time of Fiedler regularization could likely be improved with certain implementation-level optimizations to speed up the software. Performances would also likely improve if more refined grid searches or more sophisticated hyperparameter selection methods like Bayesian optimization are adopted.

\section{Discussion}
The above experiments demonstrated several points of interest. The poorer performance of L1 regularization in MNIST and CIFAR10 stands in sharp contrast to the much higher performance of Fiedler regularization, a weighted L1 penalty. It is generally acknowledged that L1 regularization does not enjoy good empirical performance in deep learning models. The precise reason why this happens not exactly known.  Previous studies have adopted a group-lasso formulation for regularization of deep NNs and have obtained good performance \citep{scardapane2017group}. These results suggest that modifications of L1 through weighting or other similar schemes can often drastically improve empirical performance. 

We have tracked the algebraic connectivity of the NNs during the training process in our experiments. In general, without any regularization, the NNs tend to become more connected during training, i.e. their Fiedler value increases. In the Fiedler regularization case, the connectivity is penalized and therefore decreases during training in a very gradual manner. Interestingly, in the L1 case, the algebraic connectivity of the NN can decrease very quickly during training, often leading to disconnection of the network very early in the training process. This is likely related to L1 regularization's uniform penalization of all weights. It is therefore difficult to choose the regularization hyperparameter for L1: if it is too small, sparsity induction might occur too slowly and we would under-regularize; if it is too big, we risk over-penalizing certain weights and lowering  the model's accuracy. One advantage of a weighted scheme such as Fiedler penalization thus lies in its ability to adaptively penalize different weights during training.

While we only considered relatively simple feedforward, fully connected neural architectures, potential extensions to more sophisticated structures are straightforward. Many convolutional NNs contain fully connected layers after the initial convolutional layers. One could easily extend Fiedler regularization to this case. In the context of ResNets, where there are skip connections, the spectral graph properties we have utilized still hold, and hence Fiedler regularization could be directly applied. The spectral graph theory setup adopted in this paper generally holds for any undirected graph. An open direction is to establish appropriate spectral graph theory for regularization of directed graphs, which would be useful in training recurrent NNs.

While Fiedler regularization leads to sparsely connected NNs in theory, in practice it often takes a higher penalty value or longer training time to achieve sparsity with Fiedler regularization. This might in part be due to the optimization method chosen. It is known that generally SGD does not efficiently induce sparsity in L1 penalized models, and certain truncated gradient methods \citep{langford2009sparse} might prove more effective in this setting. 

We remark that while Fiedler regularization emphasizes regularizing based on graphical/connectivity structure, global penalization approaches such as Dropout, L2 etc could still prove useful. One could combine the two regularization methodologies to achieve simultaneous regularization.

Lastly, we have adopted a version of spectral graph theory that considers the un-normalized (combinatorial) Laplacian $\mathbf{L} = \mathbf{D}-\mathbf{|W|}$ as well as the edge expansion of the graph. A similar theory for regularization could be developed for the normalized Laplacian $\mathbf{L}' = \mathbf{I} - \mathbf{D}^{-\frac{1}{2}}\mathbf{|W|}\mathbf{D}^{-\frac{1}{2}}$ and the conductance of the graph, after appropriately accounting for the total scale of the NN. While the combinatorial Laplacian that we considered is related to the notion of RatioCuts, the normalized Laplacian is associated with the notion of NCuts. Both notions could in theory be used for reducing connectivity/co-adaptation of the NN.

\section*{Acknowledgements}
We would like to thank Julyan Arbel for pointing out a mistake in the initial draft of this paper, now corrected.
\bibliography{Fiedler}
\bibliographystyle{icml2020}

\end{document}